\documentclass[a4paper,twoside]{article}

\usepackage{epsfig}
\usepackage{calc}
\usepackage{amssymb}
\usepackage{amstext}
\usepackage{amsmath}
\usepackage{amsthm}
\usepackage{multicol}
\usepackage{pslatex}

\usepackage{natbib}
\let\bibhang\relax

\usepackage{apalike}
\usepackage{algorithm}
\usepackage{algpseudocode}
\usepackage{subfig}
\usepackage[euler]{textgreek}
\usepackage{amssymb}
\usepackage{graphicx}
\usepackage{csquotes}
\usepackage{enumerate}
\usepackage{latexsym}
\usepackage{url}
\usepackage{subfig}
\usepackage{array}
\usepackage{spverbatim}
\usepackage{comment}
\usepackage{eurosym}
\usepackage{booktabs}
\usepackage{adjustbox}
\usepackage{subfig}
\usepackage{svg}
\usepackage{multirow}
\usepackage{adjustbox}
\usepackage{xcolor}

\usepackage{float}

\usepackage{SCITEPRESS}     

\begin{document}


\title{A Zero-Shot Classification Approach for a Word-Guessing Challenge}


\author{\authorname{Nicos Isaak\orcidAuthor{0000-0003-2353-2192}}
	\email{nicosi@acm.org}
	\affiliation{Computational Cognition Lab, Cyprus}
	
}



\abstract{The Taboo Challenge competition, a task based on the well-known Taboo game, has been proposed to stimulate research in the AI field. The challenge requires building systems able to comprehend the implied inferences between the exchanged messages of guesser and describer agents. A describer sends pre-determined hints to guessers indirectly describing cities, and guessers are required to return the matching cities implied by the hints. Climbing up the scoring ledger requires the resolving of the highest amount of cities with the smallest amount of hints in a specified time frame. Here, we present TabooLM, a language-model approach that tackles the challenge based on a zero-shot setting. We start by presenting and comparing the results of this approach with three studies from the literature. The results show that our method achieves SOTA results on the Taboo challenge, suggesting that TabooLM can guess the implied cities faster and more accurately than existing approaches.}

\onecolumn \maketitle \normalsize \setcounter{footnote}{0} \vfill

\section{Introduction}
Since the late fifties \citep{mccarthy2006proposal}, numerous studies have investigated ways to develop systems that automate or enhance basic human abilities. Based on the \citet{mccarthy2006proposal} proposal, where the term Artificial Intelligence (AI) was coined, the research community listed several topics to be tacked, such as neural nets, abstraction, NIP, etc. In this regard, various challenges have been proposed, such as the Turing test or the Winograd schema challenge, for developing systems humans can relate to and interact with. Given the above, researchers have increasingly become interested in identifying ways to endow machines with the necessary knowledge \citep{kn:michael2013machines, Isaak2016} that would allow machines to perform similar to humans. 

One of those challenges, the Taboo Challenge Competition (TCC) \citep{Rovatsos_Gromann_Bella_2018}, is concerned with the ability to resolve hints to cities, which in certain cases is argued to require the use of commonsense knowledge. The challenge refers to games between guessers and describers, where a describer sends hints that need to be resolved to city names by the guessers. Like with the traditional Taboo game, the tricky part of the challenge is that it requires the guessers to speculate the domain of the describers so that players from the same region have better chances of getting higher scores. Given the above, the challenge requires the development of diversity-aware guessers to tackle games previously played with interactive decision-making strategies by human players. According to \citet{Rovatsos_Gromann_Bella_2018}, although humans can easily solve Taboo-like games, the performance of automated approaches is still significantly lacking.

It is believed that the challenge will lead to the development of diversity-aware agents, something partially overlooked by other AI challenges. On another, the challenge allows the development of agents based on various solutions, such as Classic AI, modern AI, or blended hybrid solutions that combine the best of two worlds.

Motivated by the difficulty of having AI agents tackle Taboo Games, we introduce TabooLM, a language model approach for the Taboo Challenge competition. We start by demonstrating the challenge itself, presenting the aspects of our system architecture, and continuing with our experimental - evaluation results. To date, this is the first published work to report results on the feasibility of a zero-shot setting approach. According to our results, TabooLM outperforms the other systems in prediction accuracy, suggesting that it can guess the implied cities faster and more accurately than previously used approaches.

\section{The Taboo Challenge}
\begin{figure*}[t]
	\centerline{\includegraphics[width=\textwidth]{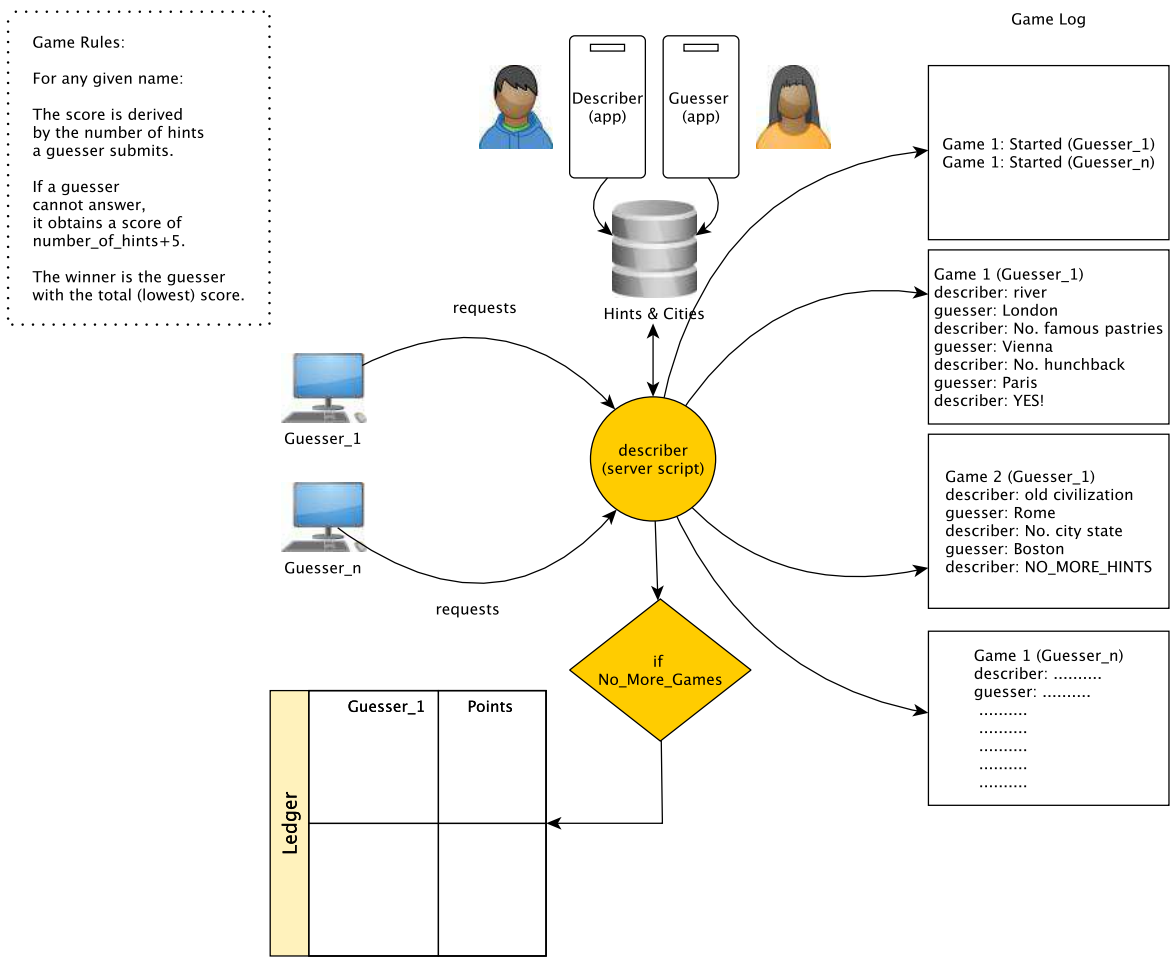}}
	\caption{Data Collection and Evaluation of the Taboo Challenge Competition}
	\label{catexample}
\end{figure*}

The challenge refers to Taboo-like games where players exchange request-response messages describing a concept without using taboo words, that is, words making the concept too easy to guess \citep{Rovatsos_Gromann_Bella_2018}. Instead of having humans, the challenge refers to a new type of game\footnote{\url{https://www.essence-network.com}} played between machines. Specifically, the challenge consists of request-response messages between describer and guesser agents trying to guess city names. To illustrate, a game starts with a guesser sending an online request to a describer (see Fig. \ref{catexample}). Next, a describer returns a hint referring to a city and waits for the guesser's response, that is, the implied city name. All subsequent responses to incorrect guesses contain the answer ``no'' and a new hint. On the other hand, the response to a correct guess is a simple yes. In case no more hints are available, the answer ``no more hints" is returned. 

The evaluation consists of running the guesser agents a predefined number of times, each time playing a different game. A guesser's score is derived by the number of guesses it submits (see game rules in Fig. \ref{catexample}). If a guesser fails to answer, it obtains a score of  number\_of\_hints+5. A guesser's total score is the sum of the individual game scores it played. The winner is the guesser with the total (lowest) score. 

According to \citet{Rovatsos_Gromann_Bella_2018}, to reduce time complexity:
\begin{itemize}
\item The domain of concepts is limited to include only popular cities, ---unknown to the guessers.
\item The hints describing cities are limited to simple noun phrases ---plus adjectives and/or adverbs.
\item Each game had to be answered in the limited time frame of twenty minutes.
\end{itemize}

\section{Related Work}
The first and only Taboo Challenge took part as a side event of IJCAI 2017 \citep{Rovatsos_Gromann_Bella_2018}. According to the organizers, 
a describer (server script) was programmed to interact with several guesser agents via an API. The hints and Taboo words were previously collected from games played by eighty-two crowd workers (based in the UK and USA). Starting from an initial set of 300 cities, they ended up with 226 cities, for which more than one worker generated eight to twelve taboo words ---cities with no more than three taboo words were eliminated. Next, via a Web and a mobile application (see Fig. \ref{catexample}), based on 226 cities, thirty native English speakers generated 283 games. Based on the collected hints, around 25\% of the games were solved right after one hint, and around 50\% using two to four hints. Finally, the challenge evaluation procedure was fully automated, and each guesser agent was tested on a subset of 109 games (see Table \ref{1sttaboo}). Prior to the competition, participants were given access to several games in order to train their guesser agents.

\citet{inproceedingsTabooIsaak} tackled the challenge through a commonsense reasoning system, originally designed for the Winograd Schema Challenge. Given a predefined list of cities, for any hint, the system searches the English Wikipedia to build semantic scenes, that is, relations between nouns, verbs, and adjectives found in English sentences. Then, it feeds the scenes to a Learner and a Reasoner, and through chaining, it outputs logical inference rules (e.g., city name, LearnerWeight). Finally, the engine responds to a describer with the city with the bigger LearnerWeight. For instance, the scene, mahal([variable:0]):- mausoleum([variable:0, exists:0]), tells us that \textit{mahal} is a mausoleum. Based on the results, the system won only six games out of 109 (5.50\%) with 197 guesses and a total score of 816.  

\citet{Dankers2017ModellingWA}, via word embeddings, they associate hints with cities. The idea behind word embeddings is that words that appear in similar contexts tend to have related meanings. To build their embeddings semantic space, they utilize and filter data from online sources such as Wikipedia, Wikivoyage, and NomadList. Basically, based on multiple game strategies and cosine similarity, they calculate and store the association between the vectors of hints and cities. In the end, they return the city with the highest score. In case a hint is not presented in the semantic space, they employ another strategy via searching pre-trained word2Vec vectors, previously trained on Google News. Based on the results, the system won thirteen games out of 109 (11.9\%) with 293 guesses and a total score of 773.

Similar to \citet{Dankers2017ModellingWA}'s approach, \citet{Koksal2017ModellingWA} tackles the challenge via similarities between hints and cities, based on the pre-trained Skip-gram Word2Vec architecture. Moreover, instead of a simplified city-level approach, they employ a different country-level approach. In short, given a list of countries, for any hint, they calculate the country-hint similarity to return a list of the top three countries. Similarly, for each city, they estimate the city-hint similarity. To calculate the similarities between cities and hints, they use a heuristic approach that increases the likelihood of an answer by multiplying the similarity factor by each city population. According to the authors, this approach returns a country's capital, or the country's most popular city. According to the results, this system was ranked 1st by winning eighteen games out of 109 (16.5\%) with 290 guesses and a total score of 745.  

\begin{table}[h]
	{\caption{Results of the First and Only Taboo Challenge.}\label{1sttaboo}} \centering
	\begin{adjustbox}{max width=\columnwidth}
\begin{tabular}{|c|c|c|c|}
	\hline 
	Team & Games Won  & Guesses & Score\tabularnewline
	\hline 
	\hline 
	\citet{Koksal2017ModellingWA} & 18 (16.5\%) & 290 & 745\tabularnewline
	\hline 
	\citet{Dankers2017ModellingWA} & 13 (11.9\%) & 293 & 773\tabularnewline
	\hline 
	\citet{inproceedingsTabooIsaak} & 06 (5.5\%) & 197 & 816\tabularnewline
	\hline 
\end{tabular}
\end{adjustbox}
\end{table}

Our work differs from previous works mainly in three key aspects. Firstly, it tackles the challenge via utilizing an out-of-the-box language model approach. Secondly, it handles the challenge as a zero-shot classification problem without training or fine-tuning. Thirdly, it does not make us of standard NLP techniques, like removing stop-words, utilizing dependency parsing, or using any kind of similarity score to enhance its decision-making mechanism. The sections below explain each of these tasks along with our system's architecture.


\section{Language Models}
Language models (LMs) refer to neural approaches that use sizeable pre-trained models, mostly fine-tuned on downstream tasks to maximize their performance. These models make it possible for machines to answer questions, write poems or music, and play games sometimes even better than humans.

Language models such as BERT \citep{devlin2018bert}, RoBERTa \citep{liu2019roberta},  GPT-2 \citep{Radford2019LanguageMA}, GPT-3 \citep{brown2020language}, DeBERTa \citep{https://doi.org/10.48550/arxiv.2006.03654}, or embeddings and architectures such as ELMO \citep{https://doi.org/10.48550/arxiv.1802.05365} and ULMFiT \citep{https://doi.org/10.48550/arxiv.1801.06146} are increasingly important to the AI research community as they revolutionized the NLP field.

Basically, these models learn the probability of word occurrence in text sequences. From this standpoint, we can have large models with billions of parameters that can be either autoregressive, like GPT-3, or able to predict sequences of sentences or a missing token from a sentence like BERT. According to the literature, the parameters of these models, an enhanced form of word embeddings, appear to store a form of knowledge that can help tackle various NLP tasks, such as question answering, pronoun resolution, text summarization, token classification, text similarity, and zero-shot classification tasks.

For instance, in a recent work, language models have been utilized to output knowledge to construct knowledge graphs \citep{https://doi.org/10.48550/arxiv.2010.11967}. For another, in the Winograd Schema Challenge (WSC), a challenging task for pronoun resolution, we can use the embedded knowledge of language models to resolve pronouns. 

All in all, language models seem to catch the relationships between words in sentences and phrases that can be used in various tasks. In case of challenging NLP problems, the parameters of these models can be further fine-tuned to downstream tasks without having to train them from scratch. In this regard, it seems that models with millions or mostly billions of parameters can tackle challenges that good-old fashion AI (GOFAI) has struggled with for many years. However, we must keep in mind that LM's computational innards are so complex that nobody understands how they really work, meaning that they are not transparent solutions ---their achievements do not seem to relate to a deeper understanding of the natural language text they are dealing with.

\subsection{Zero-shot Classification}

Zero-shot classification refers to techniques for applying language models to downstream tasks so that no further training or fine-tuning is needed. Although traditional zero-shot methods require providing some kind of descriptor to help the model predict the required task, the aim is to classify unseen classes \citep{xian2017zero,pmlr-v37-romera-paredes15,https://doi.org/10.48550/arxiv.2109.01652}. In a simplified way, zero-shot learning refers to tasks when we have annotated data for only a limited subset of our classes. Given that language-model learning refers to a specified vector space (called semantic space), zero-shot refers to the ability of these models to match unseen classes to the seen classes' vector space, which acts as a bridging source mechanism.

According to \citet{Radford2019LanguageMA}, when trained on large datasets, language models start learning various relations between text sequences without requiring explicit supervision. As a result, these models seem to have the necessary relations needed to tackle various tasks in zero-shot settings. Though not always necessary, in some cases, some instruction tuning is needed to minutely tailor the dataset to a specific zero-shot setting \cite{https://doi.org/10.48550/arxiv.2109.01652} ---e.g., large LMs like GPT-3 showed to perform better in few-shot than in zero-shot learning.

\begin{figure*}[!htb]
	\centerline{\includegraphics[width=\columnwidth]{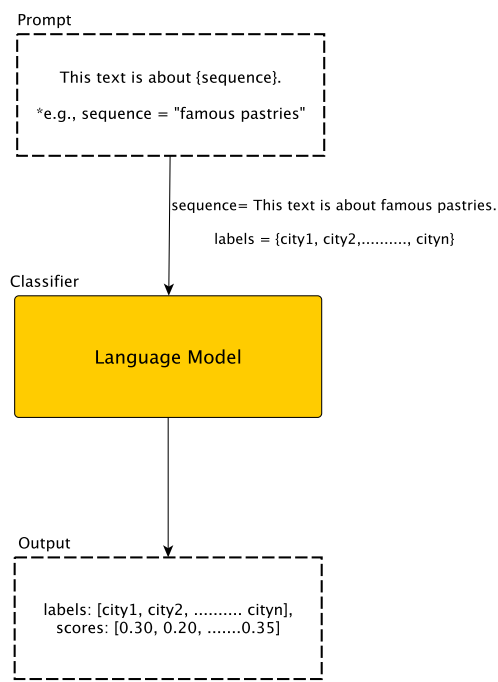}}
	\caption{Zero-Shot Classification via Language Models. The model accepts a sequence ---in some cases modified by a hypothesis\_template such us ``This text is about \textit{sequence}", followed by a list of candidate labels. The result is a list of probabilities for entailment and contradiction for each label based on the hypothesis\_template.}
	\label{tabooLM}
\end{figure*}

Nevertheless, in this paper, we approach the Taboo challenge as a natural language inference (NLI) problem \citep{maccartney2009natural}, which can be tackled via a zero-shot setting approach. Below, we will show how language models trained on datasets for determining whether a premise and a hypothesis are connected via entailment or contradiction can be utilized in a zero-shot setting to tackle the Taboo challenge (see Figure \ref{tabooLM}).

\section{System Architecture}
\label{Section: Platform Architecture}
\begin{figure*}[!htb]
	\centerline{\includegraphics[width=\textwidth]{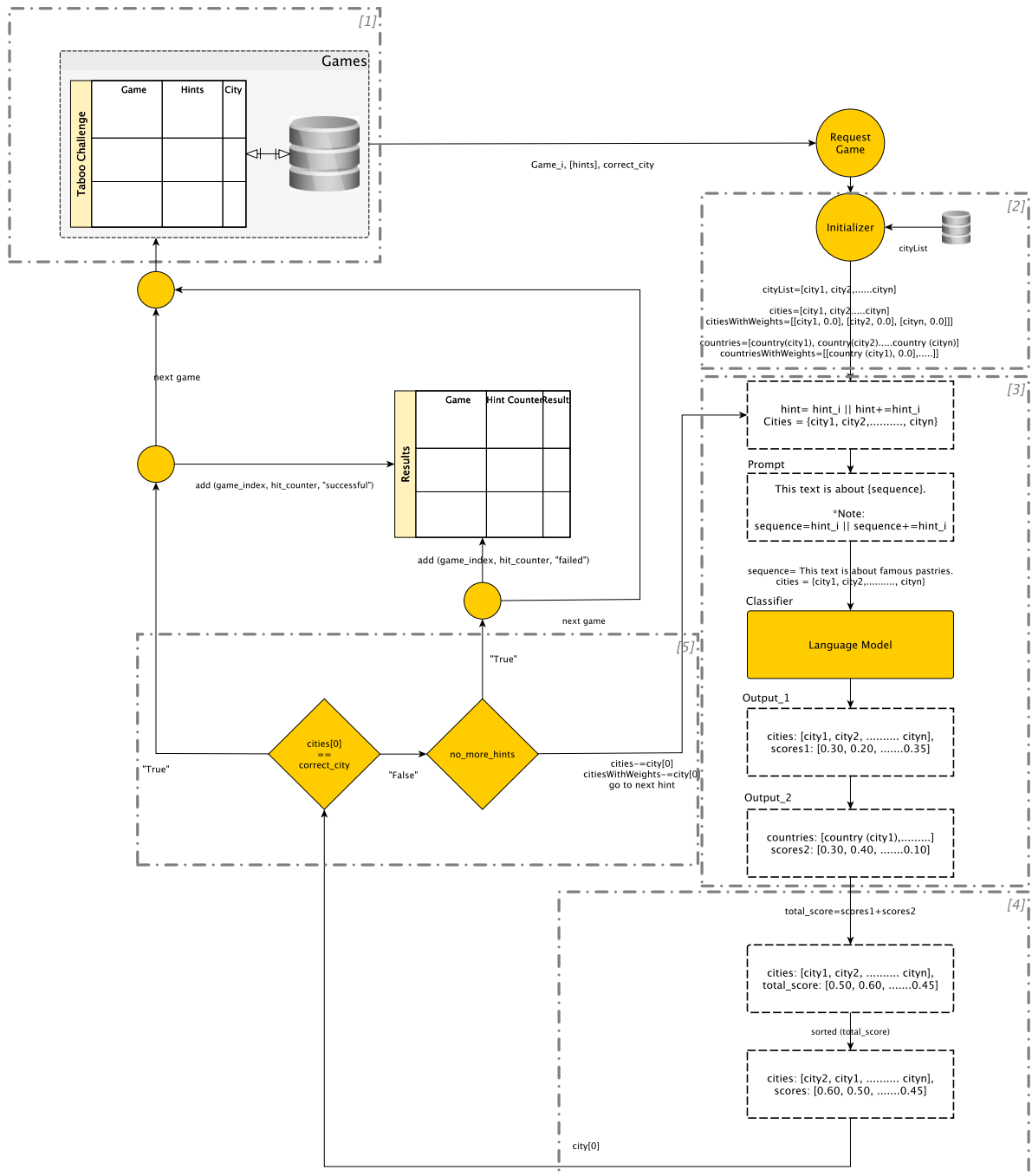}}
	\caption{Full Pipeline of TabooLM: A Zero-Shot Language Model for a Word Guessing Game. The reasoning process starts with each hint, modified by the hypothesis\_template: This text is about \textit{hint}. The various parts of the architecture are marked in blue rectangles, and are discussed in Section~\ref{Section: Platform Architecture}.}
	\label{tabooLMArchitecture}
\end{figure*}

We start by briefly discussing the main elements of our approach by presenting how it works and how it handles its semantics to guess cities given a predefined set of hints (see Figure~\ref{tabooLMArchitecture}).

Based on the constrained definition of the challenge, our system requests Taboo games from a describer, meaning lists of hints, one at a time, to return cities implied by the given hints. However, given that i) the Essence's describer service is no longer available, and ii) we only have access to the final 109 Taboo games from our previous system \citep{inproceedingsTabooIsaak}, TabooLM was designed to tackle games from a local-built service. In this regard, in its current version, TabooLM has access only to a testing set of 109 Taboo games. In this sense, from now on, when we refer to the local describer or guesser, we refer to components of our system's architecture ---although they appear to be completely different, they are, ultimately, parts of the same system.

Moreover, with respect to the first Taboo challenge, the evaluation process provided convincing evidence of a strong association between each Taboo player's biases and beliefs, according to which pairs of hint-cities were collected. In this regard, to built TabooLM we focused on utilizing large pre-trained language models\footnote{\url{https://huggingface.co}} in a zero-shot setting. From this standpoint, knowing that no training set was available, our system allows access to a broad collection of language models and handles the describer's hints in various ways.

\subsection{Loader}
At first, our guesser starts its interaction with the local describer by requesting a list of Taboo games, meaning pairs of games, hints, and correct answers (see \textit{part-1} in Figure~\ref{tabooLMArchitecture}). This interaction takes place each time no more hints are available for the current game.

\subsection{Initializer}
This component, which relates to initializing some key variables, takes part in every new round of a game. For instance, a new city list is initialized with zero weights in every new game. This is like a feature vector that, for each city-hint pair, holds the returned values of our zero-shot classification setting. Additionally, given that each city's country might improve our system's results \citep{Koksal2017ModellingWA}, the same process is repeated for each country (city)-hint pair (see part-2 in Figure~\ref{tabooLMArchitecture}).

\subsection{Zero-shot Evaluator}
Part-3 of Figure \ref{Section: Platform Architecture} illustrates how our zero-shot setting works. To instruct the model to classify cities, we start by modifying a premise which in our case is a specified sequence of text in the form of ``This text is about {hint}". The aim is to classify whether the premise entails the hypothesis, meaning a city name. For instance, given a list of hints ('tea', 'whiskey', 'kilt', 'crocodile') and a list of cities (e.g., 'Dundee', 'Athens'), the system instructs the model to predict how much the premise ``This text is about {tea}" relates to each city. In short, for each city, the system automatically runs the classifier with the same premise to output a numerical value in the range of 0-1. The classifier runs either with the flag true\_labels ``true" or ``false", where in the case of the former more than one city could be true.

Additionally, in each game, the system either runs the classifier for each hint independently or with an aggregation mechanism in the form of hint+=hint. To illustrate, in the second run ---for the second hint ``whiskey', the system either runs the premise ``This text is about {whiskey}" or combined with the first hint as ``This text is about {tea, whiskey}", against each city to output a probability list (e.g., ['Dundee', 0.05485], ['Athens', 0.0029]).

Afterward, via an API, the system loads each city's country and repeats the same procedure (e.g., ['UK', 0.30], ['Greece', 0.10], ......). In the final step, it matches cities and countries, adds their probability scores, and produces a final city list (e.g., ['Dundee', 0.35485], ['Athens', 0.1029], ......). The variability in probability values generally stems from the relation between the hypothesis and the premise values. In every game, for every new hint, the system has the option to either initialize or accumulate the weights.

In order to represent the cities according to their significance (see part-4 in Figure~\ref{tabooLMArchitecture}), the system sorts the city list in descending order with the highest probability values at the start. Once the sorting is made, it sends the first city as the answer to the given hint.

Once the city is sent, the local describer replies with a successful or an unsuccessful message (see part-5 of Figure ~\ref{tabooLMArchitecture}). In case we were able to guess the city correctly, the system adds the results to a local scoring file, in the form of \textit{game, hint, Flag}, where game and hint refer to the current game and hint indexes, and Flag to whether we were successful or unsuccessful. On the other hand, if we cannot predict the correct city, the system proceeds to the following actions: Firstly, in case no more hints are available, it adds the results to the local scoring file and moves to the next game. Secondly, if more hints are available, it proceeds to the next hint and removes the city from the current city list along with its probability value ---a wrongfully guessed city cannot be the answer to the current game.

As an example of this, consider the case of sending the city of Dundee from the final list (e.g., ['Dundee', 0.35485], ['Athens', 0.1029]). In case of an unsuccessful guess, i) the city of Dundee is removed, and ii) the system proceeds to the next hint with the left cities (e.g., ['Athens', 0.1029], ......]).

\section{Experimental Evaluation}
In this section, we present the results obtained by applying the methodology described in this paper. We describe the design along with the results of the experiments we undertook to evaluate the system performance on guessing cities from a predefined list of 109 games.

To evaluate the zero-shot performance setting, we started by investigating whether our system could produce similar or better results than previously used systems \cite{Rovatsos_Gromann_Bella_2018}. For the purposes of this experiment, the \textit{weights' initialize} Flag was enabled, meaning that with every new hint, city and country weights were initialized with zero values. Furthermore, within every game, each hint was added to the previous one in the form of hint+=next\_hint.

Our experiments ran under the DeBERTa-V3 model \citep{https://doi.org/10.48550/arxiv.2111.09543,laurerless}, fine-tuned on the multiNLI, adversarial-NLI (ANLI), fever-NLI, lingNLI, and wanli datasets. DeBERTa V3 is an enhanced version of DeBERTa, which combines BERT and RoBERTa models in an efficient novel way. In short, each model, from BERT, RoBERTa, DeBERTa, and DeBERTa V2 to DeBERTa V3, can be considered an enhanced version of the previous one. Finally, all datasets combined result in 885.242 premise-hypothesis pairs.


\subsection{Results and Discussion}
The general picture emerging from the analysis is that TabooLM correctly tackled 53 games out of 109, achieving a success rate of 49\% (see Table \ref{tabooLMresutls}). A cursory glance at Table \ref{tabooLMresutls} reveals that our approach significantly outperformed all previously used systems. This is in line with recent results where language models significantly outperformed other methods in various NLP tasks \cite{brown2020language}.

A more detailed analysis of our results shows that TabooLM not only tackled more games but it also achieved the best score of 417 points ---recall that less is more. Specifically, it outperformed \citet{Koksal2017ModellingWA}'s system by 330 points, \citet{Dankers2017ModellingWA}'s by 358 points, and finally, our previous approach by 301 points \citep{inproceedingsTabooIsaak}.
An interesting finding was that it achieved the best score with the minimum number of 267 guesses, which is very important considering the challenge difficulties (see Table \ref{tabooLMresutls}). Moreover, compared to our previous work, which due to its reasoning engine, was unable to answer all the games as it timed out frequently \citep{Rovatsos_Gromann_Bella_2018}, TabooLM tackled 109 games in 25 minutes ---experiments ran under an NVIDIA Tesla K80 GPU. Specifically, compared to an average time of 20 minutes for each game, TabooLM was found to be 77\% faster than our previously used approach.

These findings are less surprising if we consider not only the advances of deep learning approaches in the NLP field but also recent work showing ways to utilize LMs to output semantic relations to help tackle NLP tasks \citep{https://doi.org/10.48550/arxiv.2010.11967}. It seems that LMs, indeed, could be utilized in zero-shot settings to achieve state-of-the-art results \citep{Radford2019LanguageMA}.

Further experiments we undertook revealed that the capacity of the language model in our zero-shot setting relates to its training size. For instance, we further analyzed the relationship between models trained on different datasets and their success in the Taboo challenge competition. The data provide convincing evidence of a link between accuracy and the variety of the training datasets. In short, it seems that as the training data increases, we can get better semantics to do a better word guessing. This is in line with other work in which leveraging larger language models or training with larger datasets improve system performance \citep{Radford2019LanguageMA,Isaak2017HowTA}. A cursory look at Table \ref{tabooLMresutlsTraining} reveals that as the number of training datasets increases, the number of unanswered games decreases. It seems that the increase in accuracy was due to the increase in the number of training datasets, meaning that a variety of training datasets limits the situations where a hint-city relation is unlike anything an LM has met in the training phase (see Table \ref{tabooLMresutlsTraining}).

\begin{table}[h]
	{\caption{Results of TabooLM Compared to Systems Participated in the First Taboo Challenge.) }\label{tabooLMresutls}} \centering
	\begin{adjustbox}{max width=\columnwidth}
		\begin{tabular}{|c|c|c|c|}
			\hline 
			Team & Games Won  & Guesses & Score\tabularnewline
			\hline 
			\hline 
			TabooLM & 53 (49\%) & 267 & 417\tabularnewline
			\hline 
			\citet{Koksal2017ModellingWA} & 18 (16.5\%) & 290 & 745\tabularnewline
			\hline 
			\citet{Dankers2017ModellingWA} & 13 (11.9\%) & 293 & 773\tabularnewline
			\hline 
			\citet{inproceedingsTabooIsaak} & 06 (5.5\%) & 197 & 816\tabularnewline
			\hline 
		\end{tabular}
	\end{adjustbox}
\end{table}

\begin{table}[h]
	{\caption{Results of TabooLM Based on a Various Models. Results show that a variety of training datasets leads to better guessing.) }\label{tabooLMresutlsTraining}} \centering
	\begin{adjustbox}{max width=\columnwidth}
		\begin{tabular}{|c|c|}
			\hline 
			Model & Games Won \tabularnewline
			\hline 
			\hline 
			facebook/bart-large-mnli & 27\tabularnewline
			\hline 
			bart-large-mnli-yahoo-answers & 38\tabularnewline
			\hline 
			DeBERTa\_v3\_large\_mnli-\_fever\_anli\_ling-\_wanli & 53\tabularnewline
			\hline 
		\end{tabular}
	\end{adjustbox}
\end{table}




\section{Conclusion}
We have shown TabooLM, a system that takes queries in a city-hint format and utilizes language models in a zero-shot setting to return ranked lists of cities implied by the given hints. Given a list of hints, it iterates from top to bottom and matches those hints with popular cities worldwide. Although it was built explicitly for the Taboo challenge competition, the system can be used with any task involving a word-guessing problem.

Compared to previous work, the results provide convincing evidence that our system can achieve state-of-the-art results.
In this regard, the results suggest that solutions utilizing language models in a zero-shot setting can be used to tackle challenging NLP tasks. However, given that the computational innards of these kinds of models are complex, further gains could be achieved via transparent solutions that employ additional semantic analysis of city-hint pairs. 

Future studies could blend both modern and classic AI in order to build transparent hybrid solutions. Among possible directions, systems that construct the building of knowledge graphs from language models could offer a better solution.

\bibliographystyle{apalike}
{\small
	\bibliography{isaak}}

\end{document}